%% file: main.tex
\documentclass[letterpaper]{article} 
\usepackage{aaai2026}  

\input{preamble}
\input{macros}

\usepackage{times}  
\usepackage{helvet}  
\usepackage{courier}  
\usepackage[hyphens]{url}  
\usepackage{graphicx} 
\usepackage{natbib}  
\usepackage{caption} 
\title{\nsvqa: Grounding Long-Form Video Understanding in Temporal Logic and Neuro-Symbolic Reasoning}

\author {
    Sahil Shah\textsuperscript{\rm 1},
    S P Sharan\equalcontrib\textsuperscript{\rm 1},
    Harsh Goel\equalcontrib\textsuperscript{\rm 1},
    Minkyu Choi\textsuperscript{\rm 1},
    Mustafa Munir\textsuperscript{\rm 1},\\
    Manvik Pasula\textsuperscript{\rm 2},
    Radu Marculescu\textsuperscript{\rm 1},
    Sandeep Chinchali\textsuperscript{\rm 1}
}
\affiliations {
    The University of Texas at Austin, Texas, USA\textsuperscript{\rm 1}\\
    Independent Researcher, USA\textsuperscript{\rm 2}\\
    \{ss96869, spsharan, harshg99, minkyu.choi, mmunir\}@utexas.edu\\
}
\nocopyright

\usepackage{bibentry}

\begin{document}


\maketitle



\begin{abstract}
While vision-language models (VLMs) excel at tasks involving single images or short videos, they still struggle with Long Video Question Answering (LVQA) due to its demand for complex multi-step temporal reasoning.
Vanilla approaches, which simply sample frames uniformly and feed them to a VLM along with the question, incur significant token overhead. This forces aggressive downsampling of long videos, causing models to miss fine-grained visual structure, subtle event transitions, and key temporal cues.
Recent works attempt to overcome these limitations through heuristic approaches; however, they lack explicit mechanisms for encoding temporal relationships and fail to provide any formal guarantees that the sampled context actually encodes the compositional or causal logic required by the question. 
To address these foundational gaps, we introduce \nsvqa, a training-free, plug-and-play neuro-symbolic pipeline for LVQA. \nsvqa\ first translates a natural language question into a logic specification that models the temporal relationship between frame-level events. Next, we construct a video automaton to model the video's frame-by-frame event progression, and finally employ model checking to compare the automaton against the specification to identify all video segments that satisfy the question's logical requirements.
Only these logic-verified segments are submitted to the VLM, thus improving interpretability, reducing hallucinations, and enabling compositional reasoning without modifying or fine-tuning the model. 
Experiments on the LongVideoBench and CinePile LVQA benchmarks show that \nsvqa\ significantly improves performance by \textbf{over} $\mathbf{10\%}$, particularly on questions involving event ordering, causality, and multi-step reasoning. We open-source our code at https://utaustin-swarmlab.github.io/NeuS-QA/.
\end{abstract}

\section{Introduction}

\begin{figure}[t!]
    \centering
    \begin{subfigure}[b]{\linewidth}
        \centering
        \includegraphics[width=\linewidth, clip, trim=0cm 14.5cm 43.3cm 0cm]{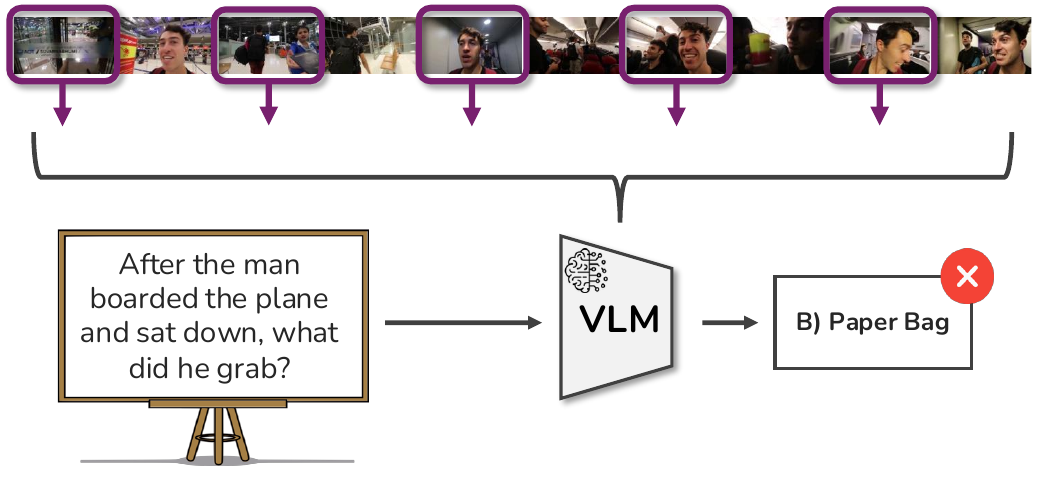}
        \caption{\textbf{Vanilla VLM Prompting} uniformly samples frames across the video regardless of relevance, often missing key temporal context.}
        \label{fig:teaser:a}
    \end{subfigure}
    
    \vspace{0.5em}
    
    \begin{subfigure}[b]{\linewidth}
        \centering
        \includegraphics[width=\linewidth, clip, trim=0cm 14.5cm 43cm 0cm]{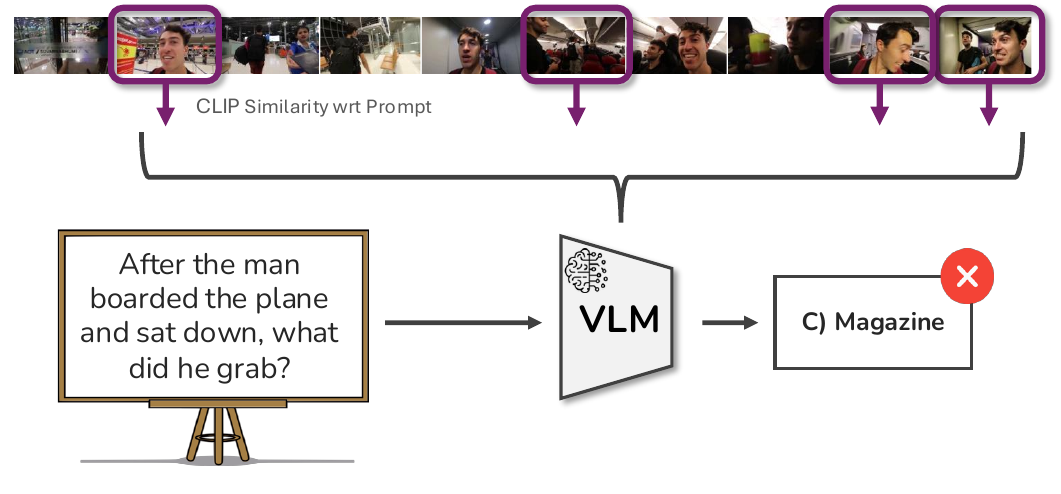}
        \caption{\textbf{Heuristic Retrieval} ranks frames based on semantic similarity to the query but lacks temporal structure and formal guarantees.}
        \label{fig:teaser:b}
    \end{subfigure}
    
    \vspace{0.5em}
    
    \begin{subfigure}[b]{\linewidth}
        \centering
        \includegraphics[width=\linewidth, clip, trim=0cm 5.5cm 43cm 0cm]{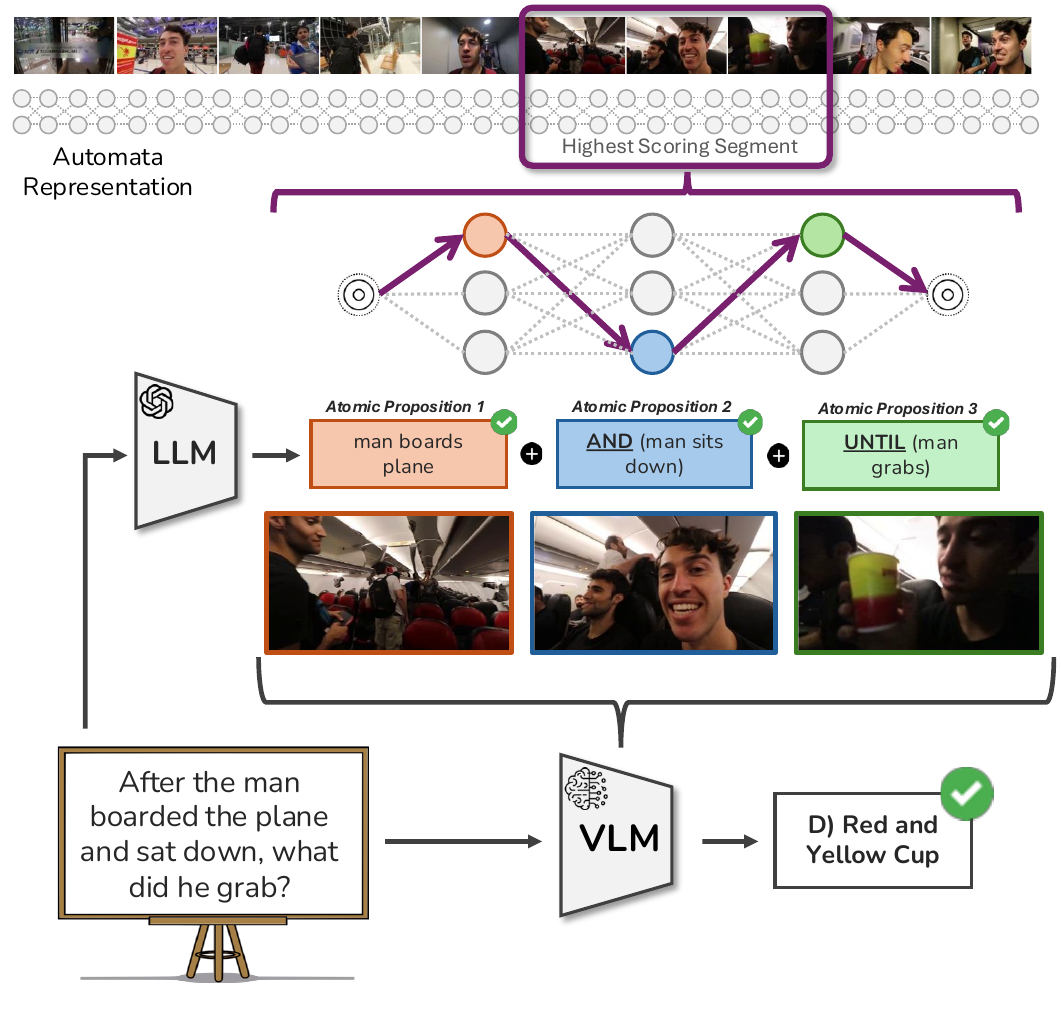}
        \caption{\textbf{Ours (\nsvqa)} builds a video automaton over grounded propositions, constructs a temporal logic specification from the query, and uses model checking to retrieve logic-satisfying video segments leading to more accurate and interpretable answers.}
        \label{fig:teaser:c}
    \end{subfigure}
    
    \caption{Comparison of frame selection strategies to answer \textit{temporally-grounded} questions over \textit{long-form} videos.}
    \label{fig:teaser}
    \vspace{-3ex}
\end{figure}

    Recent advances in video-language models (VLMs) have led to significant progress in visual question answering (VQA) for static images and short video clips \citep{park2024too, wang2024videoagent, ye2025re, wang2025videotree, choudhury2023zero}. However, as real-world applications shift toward longer-form content such as surveillance streams, egocentric vlogs, and movie scenes, users are increasingly asking more complex, temporally grounded questions that span multiple events \citep{wu2024longvideobench, rawal2024cinepile, choi2024towards, wang2024lvbench, tan2025allvb}. For example, as illustrated in \Cref{fig:teaser}, a user watching a one-hour travel vlog might ask:
    \textit{``After the man boarded the plane and sat down, what did he grab?"}
    
    Answering such questions demands more than surface-level perception; it requires \ding{202} \textbf{semantic grounding} to \textit{identify} entities and their interactions, \ding{203} \textbf{temporal reasoning} to \textit{track} and \textit{interpret} the sequence of events, and \ding{204} \textbf{compositionality} to \textit{integrate} these temporally distinct sub-events into a coherent reasoning chain
    that aligns with the structure and intent of the query.

    A common strategy in video question answering with VLMs is to uniformly sample frames across the video. While this method is sufficient for short clips, it becomes problematic as video length increases: the volume of visual input grows rapidly, forcing models to either exceed token limits or aggressively downsample videos. In doing so, they risk skipping large portions of the video and missing \textit{fine-grained visual details}, \textit{subtle event transitions}, and \textit{key temporal cues} crucial for answering complex temporal queries. 
        
    Long-form videos often follow a \textit{narrative} arc, with multiple scenes unfolding sequentially; yet many questions about them pertain to only a specific portion of that story. \Cref{fig:teaser} illustrates a vacation vlog that depicts a man's full journey from entering the airport to de-boarding to finally checking into a hotel room. But the user's query requires information solely from the in-flight segment. This is a common pattern in long-form video question answering (LVQA): although the content is broad, most queries target specific temporal windows.
    
    Therefore, a \textit{natural alternative} to uniform sampling is to direct the model's attention to a small, targeted segment of the video that is \underline{most relevant} to the query. However, it raises a core challenge: \textit{How can we retrieve the correct segment that is semantically and temporally aligned with the question, without already knowing the answer?}
        
    To address this, we argue that temporal logic (TL) offers a powerful solution. It allows us to formalize \textit{how} events unfold over time by symbolically representing their structure using operators such as ``until," ``eventually," and ``always." This makes it possible to precisely specify the kinds of temporal patterns a question refers to (like an event occurring after another), and to systematically filter for video segments that satisfy these constraints. 
    
    To this end, we propose \nsvqa, a neuro-symbolic pipeline that combines the structure of TL with the perceptual capabilities of VLMs to accurately answer \textit{temporally complex questions over long-form videos}. Given a question in natural language, \nsvqa\ first translates it into a TL specification that encodes the desired sequence of events. From the video, it then constructs a video automaton by assigning scores for semantic propositions in individual frames, and applies model checking to identify video segments that satisfy this specification. Finally, a VLM is queried on these verified segments, enabling precise, temporally grounded answering without exposing the model to irrelevant or misleading context.


\section{Related Works}
\label{sec:related-works}
    
    \paragraph{Video Question Answering.}
        
    Early VQA systems \citep{antol2015vqa} typically relied on fully supervised architectures that encoded entire videos using convolutional and recurrent networks, encoder-decoder pipelines, or attention mechanisms over video features \citep{zhu2016visual7w, ye2017video, jang2017tgif}. However, these models lacked scalability to longer videos and were constrained by limited visual-linguistic reasoning. The emergence of VLMs \citep{bai2025qwen2, internvl2024v2, achiam2023gpt, li2024llava, li2024aria} has enabled zero-shot VQA across diverse domains with stronger generalization. However, these models typically operate by consuming a fixed set of frames sampled from the entire video, limiting their ability to handle long-form videos with complex temporal structures. Their reliance on global frame pooling or simple attention mechanisms often leads to irrelevant or diluted context when presented with multi-event sequences \citep{wu2024longvideobench, wang2024lvbench}, resulting in degraded performance.
    
     To address the limitations of global frame sampling, recent state-of-the-art (SOTA) approaches adopt retrieval-augmented strategies that identify and condition on the most relevant video segments \citep{park2024too, wang2024videoagent, ye2025re, song2024moviechat+, wang2025videotree, islam2025bimba, choudhury2023zero}. While effective at reducing irrelevant context, these methods often rely on coarse heuristics or textual summarization, which sacrifice visual fidelity and lack formal structure. In contrast, \nsvqa\ uses temporal logic to guide frame selection, yielding a more interpretable and precise grounding of video content.

    \paragraph{Symbolic Representations for Video.}
    Symbolic representations have been explored across fields such as robotics~\citep{puranic2021learning,hasanbeig2019reinforcement,shoukry2017linear}, autonomous vehicles~\citep{zheng2025neurostrata, mehdipour2023formal}, generative models~\citep{shah2025challenge, sharan2025neuro, choi2025we}, and structured neural networks~\citep{manginas2024nesya, hashemi2023neurosymbolic}. In the context of vision, symbolic methods have been employed to support long-form understanding~\citep{long1, long2, long3, long-form-video}; however, these typically rely on latent-space embeddings, sacrificing interpretability. \citet{yang2023specification} and \citet{choi2024towards} move toward representing videos as formal models to enable structured reasoning. \nsvqa~follows a similar philosophy, modeling videos as verifiable automata to enable interpretable and temporally grounded analysis.

\section{Preliminaries}

    To illustrate our approach, we begin by introducing a running example. Suppose, given a video, we ask \textit{``After the woman pours hot water over granola and spoons yogurt into the bowl, what does she place as a topping?"}

    \paragraph{Temporal Logic.}
    Temporal Logic (TL) is a formal language used to describe the progression of events through a combination of logical and temporal operators. In TL, individual events are represented as atomic propositions, statements that evaluate to either \texttt{True} or \texttt{False}. These propositions are combined using logical operators such as AND (\andlogic), OR (\orlogic), NOT (\notlogic), and IMPLY (\imply), as well as temporal operators like ALWAYS (\always), EVENTUALLY (\eventually), NEXT (\ournext), and UNTIL (\until).
    
    Returning to our running example, we define a set of atomic propositions $\propset$ and a corresponding TL specification $\spec$ as follows:
        \begin{equation*}
            \label{eq:eg_spec}
            \begin{aligned}
                \propset &= \{\, \text{woman pours hot water over granola}, \\
                         &\quad \text{woman spoons yogurt into bowl}, \\
                         &\quad \text{woman places topping} \,\}, \\
                \spec    &= (\text{woman pours hot water over granola} \ \wedge \\
                         &\quad \text{woman spoons yogurt into bowl}) \ \wedge \ \diamondsuit \\
                         &\quad\text{woman places topping}.
            \end{aligned}
        \end{equation*}
    
    This specification encodes the query where the woman first pours hot water over granola, \textit{and} adds yogurt to the bowl, and \textit{eventually} adds some topping at the end.
    
    \begin{figure}[t]
        \centering
        \includegraphics[width=\linewidth,clip, trim=0cm 6cm 5.5cm 0cm]{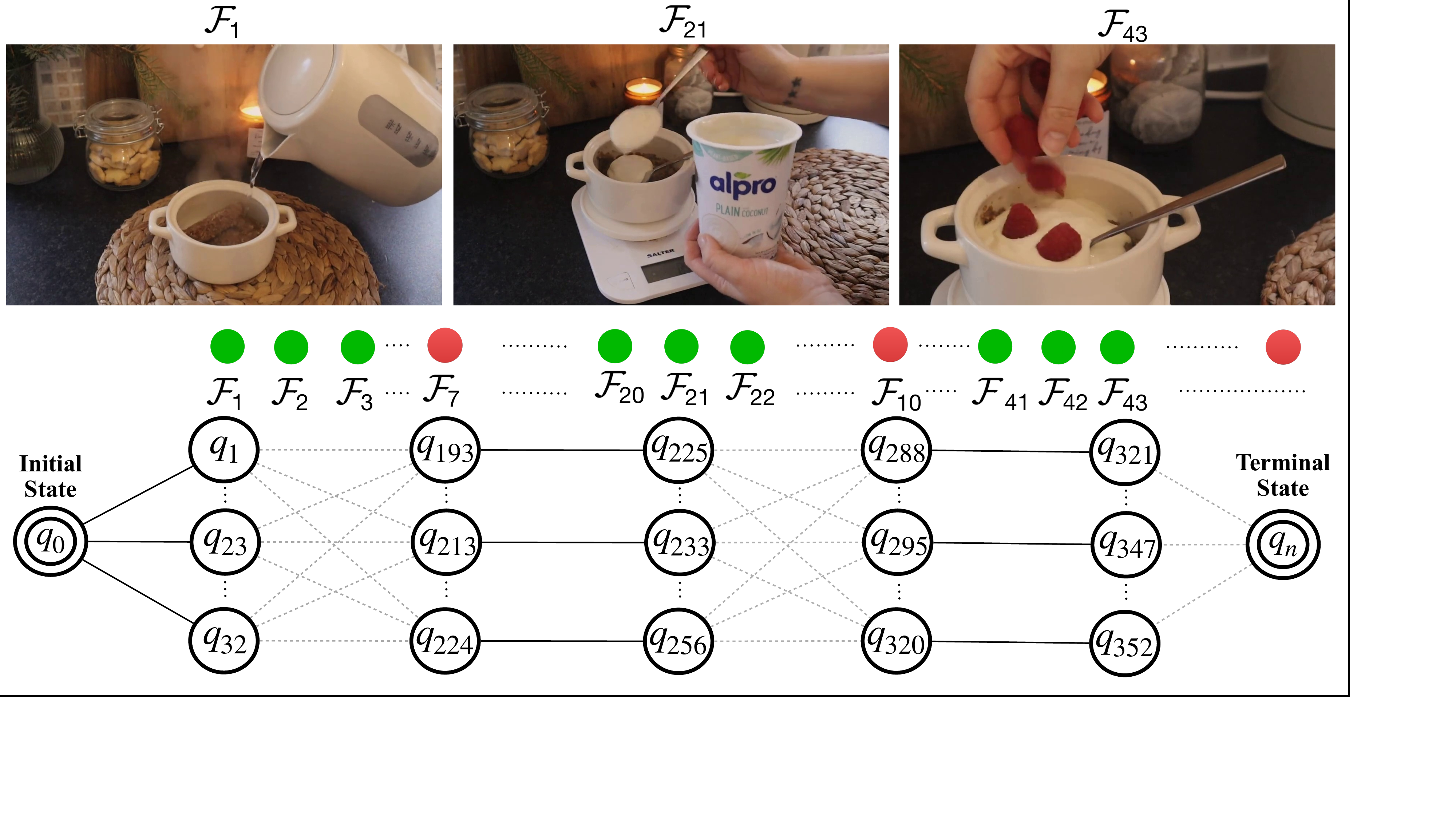}
        \caption{\textbf{Video automaton for the running example}.  
        Top: representative frames $\mathcal{F}_1$ (\emph{woman pours hot water over granola}), $\mathcal{F}_{21}$ (\emph{woman spoons yogurt into bowl}), and $\mathcal{F}_{43}$ (\emph{woman places topping}). Bottom: finite-state automaton synthesized from the TL specification in $\spec$ of the running example. \textbf{Green circles} indicate \emph{valid} frames---those in which at least one domain-specific atomic proposition (e.g., ``woman pours hot water over granola'') is detected and whose inclusion preserves all ordering constraints accumulated \textit{so far}. 
        \textbf{Red circles} mark frames that (i) contain no relevant propositions or (ii) would force a transition contradicting the specification (for instance, ``woman grabs spoon but doesn't spoon yogurt into bowl''). 
        We verify the automaton incrementally: whenever a new frame is labelled green, we extend the current run $\rho = q_0 \rightarrow \dots \rightarrow q_t$ and immediately re-evaluate the property $\rho \models \varphi$.  If the property still holds, the frame remains green; otherwise, it is re-labelled red and excluded. The procedure terminates once the accepting state $q_n$ is reached, guaranteeing that the sequence of green frames is a witness trace satisfying the specification.}
        \label{fig:running_example_dtmc}
        \vspace{-2ex}
    \end{figure}
    
    \paragraph{Video Automaton.}
    We represent a video as a stochastic finite-state automaton where each state corresponds to a sampled frame, and transitions capture temporal progression. As videos are a finite sequence of discrete frames with inherent temporal structure, we model them as a discrete-time Markov chain \citep{norris1998markov, kemeny1960finite}.
    
    Formally, the video automaton $\auto_\video$ is defined as the tuple:
    \begin{equation}
        \auto_\video = (Q, q_0, \delta, \lambda),
    \end{equation}
    where $Q$ is a finite set of states, $q_0 \in Q$ is the initial state, $\delta : Q \times Q \rightarrow [0,1]$ is a transition function such that $\delta(q_i, q_j)$ gives the probability of transitioning from state $q_i$ to state $q_j$, and $\lambda : Q \rightarrow 2^{|\propset|}$ is a labeling function that assigns a set of atomic propositions to each state. This structure allows TL formulas to be verified over the video by treating sequences of frames as labeled paths.

    \paragraph{Formal Verification.}
    \label{sec:tl}
    With formal verification, we assess whether a video automaton satisfies a given TL specification \citep{clarke1999model, huth2004logic}. We define a path $\pi$ through the automaton $\auto_\video$ as an infinite sequence of states $\pi = q_0 q_1 q_2 \dots$ such that $\delta(q_i, q_{i+1}) > 0$ for all $i \geq 0$. Each path induces a trace---the sequence of label sets assigned to the states along the path, denoted as $\text{trace}(\pi) = \lambda(q_0)\lambda(q_1)\lambda(q_2) \dots$. This trace captures the progression of observable propositions over time, as illustrated in \Cref{fig:running_example_dtmc}. We apply probabilistic model checking \citep{Baier2008} to compute the satisfaction probability $\mathbb{P}[\video \models \spec]$, quantifying how well the video trace satisfies the TL specification. This provides a formal and interpretable framework for identifying event sequences in video.

\section{Methodology}

    We present an overview of \nsvqa, which extracts temporally grounded scenes of interest from the video based on a natural language query to guide vision-language models. \nsvqa\ first identifies the key events and semantics necessary to answer the question. These are then formalized as a temporal logic specification that encodes the expected structure and ordering of events. To evaluate which parts of the video satisfy this specification, we construct an automaton whose states represent frame-level propositions predicted by a VLM. As the VLM iteratively processes consecutive video segments, we label frames with relevant propositions and incrementally build the automaton. Formal verification is then applied over this automaton to identify sequences that satisfy the query, yielding the final set of frames of interest. We outline this process in Algorithm~\ref{alg:nsvqa}.
    
    \paragraph{$\puls$: Question to Temporal Logic.}
    $\puls$ provides a structured two-shot framework for translating natural language questions into TL specifications using prompting strategies that capture both spatial and temporal relationships. Given a question $\textbf{q}$, $\puls$ extracts a set of atomic propositions $\propset = \{\prop_1, \prop_2, \ldots, \prop_n\}$ corresponding to key objects, actions, and relationships in the query. This extraction is guided by optimized prompts with carefully selected few-shot examples to ensure semantic and event-level coverage.
    
    After the propositions are identified, $\puls$ constructs a TL specification $\spec$ by mapping the temporal structure of the query onto formal operators, linking the extracted propositions accordingly. The resulting propositions and specification are detailed in \Cref{sec:tl}.

    \vspace{1ex}

{
\begingroup
\removelatexerror
\resizebox{0.9\linewidth}{!}{%
    \begin{algorithm}[H] 
        \footnotesize
        \DontPrintSemicolon
        \SetKwInOut{KwRequire}{Require}
        \SetKwInOut{KwInput}{Input}
        \SetKwInOut{KwOutput}{Output}
        
        \KwRequire{Proposition extractor $\mathcal{L}_\texttt{Q2TL}(\cdot)$, frame-level detector $\mathcal{M}_\texttt{prop}(\cdot)$, frame step size $\kappa$, automaton builder $\Theta(\cdot)$, probabilistic model checker $\mu(\cdot)$, smoothing function $\widehat{F}(\cdot)$, temporal expander $\mathcal{L}_\texttt{ext}(\cdot)$, trimming function $f_\texttt{trim}$, vision-language answerer $\mathcal{M}_\texttt{VQA}(\cdot)$, satisfaction threshold $\tau$}
        \KwInput{Video $\video$, question $\mathbf{q}$, answer choices $\mathcal{C}$, ground-truth answer $c^\ast$}
        \KwOutput{Boolean result $y \in \{\texttt{True}, \texttt{False}\}$}
        
        $\propset, \spec \leftarrow \mathcal{L}_\texttt{Q2TL}(\mathbf{q})$ \tcp*{Convert question to TL}
    
        $Z \leftarrow \emptyset$ \tcp*{Create detection matrix}
            
        $\auto_{\video,0} \leftarrow \Theta(\emptyset, \propset, Z)$ \tcp*{Initialize automaton}

        \For{$t \gets 0$ \KwTo $\mathrm{length}(\video) - \kappa$}{
            $\mathcal{F}_t \gets \{\video_t, \ldots \video_{t+\kappa} \}$ \tcp*{Select sequence of frames}
            
            \For{$\prop_i \in \propset$}{
                $Z_{t, \prop_i} \leftarrow \mathcal{M}_\texttt{prop}(\prop_i, \mathcal{F}_t)$ \tcp*{Find frames by proposition}
            }
            $\auto_{\video,t} \leftarrow \Theta(\auto_{\video,t}, \propset, Z_{0:t})$ \tcp{Extend automaton}
            
            $\mathbb{P}[\auto_{\video,t} \models \spec] \leftarrow \mu(\video, \spec)$ \tcp*{Compute satisfaction probability}
            
            \If{$\widehat{F}(\mathbb{P}[\auto_{\video,t} \models \spec]) > \tau$}{ \tcp*{Calibrated probability is above threshold}
                \textbf{break}
            }
        }
        
        $(t_{\text{start}}, t_{\text{end}}) \leftarrow \video_{[\texttt{satisfy}]}$ \tcp*{Obtain automaton frames of interest}
        
        $(\alpha, \beta) \leftarrow \mathcal{L}_\texttt{ext}(\mathbf{q})$ \tcp*{Find extension spans}
        
        $\video' \leftarrow f_\texttt{trim}(\video, t_{\text{start}} + \alpha, t_{\text{end}} + \beta)$ \tcp*{Trim to extended frames of interest}
        
        $\hat{c} \leftarrow \mathcal{M}_\texttt{VQA}(\mathbf{q}, \mathcal{C}, \video')$ \tcp*{compute VQA}
        $y \leftarrow (\hat{c} = c^\ast)$ \tcp*{determine match}
        
        \caption{\nsvqa\ Core Logic}
        \label{alg:nsvqa}
    \end{algorithm}
}
\endgroup
}

        \begin{table*}[t]
        \centering
        \resizebox{\linewidth}{!}{
            \begin{tabular}{clcccccccc}
                \toprule
                \multicolumn{2}{c}{\multirow{2}{*}{\textbf{Method}}} & \multirow{2}{*}{\textbf{Frames}}                          & \multicolumn{4}{c}{\textbf{Category}} & \multirow{2}{*}{\textbf{Overall ($\uparrow$)}}                                                                       \\
                \cmidrule(lr){4-7}
                                                                     &                                                           &                                       & T3E                                            & E3E            & T3O            & O3O            &                & \\
                \midrule
                \multirow{6}{*}{\shortstack[c]{Vision Foundation                                                                                                                                                                                                                                \\Models}}
                                                                     & LLaVA-OneVision-Qwen2-7B \citep{li2024llava}              & 32                                    & 43.40                                          & 64.86          & 54.39          & 43.18          & 53.07            \\
                                                                     & Aria \citep{li2024aria}                                   & 64                                    & 52.83                                          & 58.11          & 59.65          & 36.36          & 53.07            \\
                                                                     & Qwen2.5-VL-7B-Instruct \citep{bai2025qwen2}               & 32                                    & 39.62                                          & 58.11          & 54.39          & 45.46          & 50.44            \\
                                                                     & GPT-4o-2024-08-06 \citep{achiam2023gpt}                   & 32                                    & 43.40                                          & 58.11          & 47.37          & 45.45          & 49.56            \\
                                                                     & Qwen2-VL-7B-Instruct \citep{yang2024qwen2technicalreport} & 32                                    & 45.28                                          & 54.05          & 45.61          & 47.73          & 48.68            \\
                                                                     & InternVL2-8B \citep{internvl2024v2}                       & 32                                    & 39.62                                          & 48.65          & 49.12          & 43.18          & 45.61            \\
                \midrule
                \multirow{4}{*}{\shortstack[c]{Structured Reasoning                                                                                                                                                                                                                             \\Frameworks}}
                                                                     & VideoTree \citep{wang2025videotree}                       & 8-32                                  & \textbf{56.38}                                 & 50.00          & 50.00          & 43.84          & 50.47            \\
                                                                     & LVNet \citep{park2024too}                                 & 12                                    & 41.94                                          & 54.17          & 35.48          & 46.15          & 45.59            \\
                                                                     & VideoAgent \citep{wang2024videoagent}                     & 32                                    & 45.45                                          & 52.63          & 27.27          & 27.27          & 40.38            \\
                                                                     & T* \citep{ye2025re}                                       & 8                                     & 28.95                                          & 35.56          & 48.33          & 22.22          & 35.75            \\
                \midrule
                \multirow{4}{*}{\shortstack[c]{\nsvqa\ (Ours)}}
                                                                     & \nsvqa\ + Qwen2.5-VL-7B-Instruct                          & 32                                    & 54.72                                          & 62.16          & \textbf{66.67} & \textbf{54.55} & \textbf{60.09}   \\
                                                                     & \nsvqa\ + Aria                                            & 32                                    & 54.72                                          & 64.86          & 63.16          & 47.73          & 58.77            \\
                                                                     & \nsvqa\ + LLaVA-OneVision-Qwen2-7B-ov                     & 32                                    & 50.94                                          & \textbf{70.27} & 56.14          & 52.27          & 58.77            \\
                                                                     & \nsvqa\ + GPT-4o-2024-08-06                               & 32                                    & 49.06                                          & 68.92          & 59.65          & 50.00          & 58.33            \\
                \bottomrule
            \end{tabular}
        }
        \caption{\textbf{Benchmarking Accuracy (\%) of SOTA VQA Models.} \nsvqa, when used alongside different VLM backbones, outperforms raw foundation models and structured reasoning frameworks on LongVideoBench.}
        \label{tab:maintable}
        \vspace{-2ex}
    \end{table*}

    \paragraph{Automaton Representation of Video.}
    Given a TL specification $\spec$ from $\puls$, we construct a formal representation of the video by incrementally building a discrete-time Markov chain automaton $\va$ grounded in event-based atomic propositions. Starting from an empty automaton $\auto_{\video,0}$ (\texttt{L3}), we process the video in frame windows $\mathcal{F}_t$ of size $\kappa$ (\texttt{L5}), building a probabilistic state-based model that captures the evolution of identified events over time.
    
    For each frame sequence $\mathcal{F}_t$, the system computes probabilistic detections $Z_{t, p_i} \in [0, 1]$, indicating the likelihood that a proposition $p_i \in \propset$ is present, using a vision-language model $\mathcal{M}_{\text{prop}}(p_i, \mathcal{F}_t)$ (\texttt{L7}). These values are derived from the model’s \texttt{Yes}/\texttt{No} logits, calibrated using a held-out validation set (see Appendix) to improve the reliability of semantic assessments. Based on the cumulative detections $Z_{0:t}$ and the current automaton state $\auto_{\video,t-1}$, we update the automaton via the transition builder $\Theta(\auto_{\video,t}, \propset, Z_{0:t})$ (\texttt{L8}). This automaton incrementally models the temporal progression of event-based propositions throughout the video.
    
    \paragraph{Formal Verification for Scenes of Interest.}
    We use formal verification to identify all time intervals where the TL specification $\spec$ is satisfied. This process employs established model-checking techniques to enable rigorous temporal analysis over the video automaton (\texttt{L9}).
    
    At each timestep $t$, given the current automaton $\auto_{\video,t}$, we compute the satisfaction probability as:
    \begin{equation}
    \mathbb{P}_t = \mathbb{P}[\auto_{\video,t} \models \spec] = \mu(\video, \spec).
    \end{equation}
    This satisfaction probability is computed using probabilistic model checking, implemented via the STORM model checker \citep{storm,stormpy}. This process operates under the Probabilistic Computation Tree Logic (PCTL) framework, where $\mu(\cdot)$ analyzes the state labels and transition probabilities in $\auto_{\video,t}$ to assess how likely the specification $\spec$ is satisfied.
    
    To obtain a stable and calibrated signal, we apply a smoothing function $\widehat{F}(\cdot)$ that maps raw satisfaction probabilities into per-frame detection scores, defined as:
    \begin{equation}
    \widehat{F}(c) = \frac{1}{1 + e^{-\gamma (c - \tau)}},
    \end{equation}
    where $c$ is the raw satisfaction probability, $\tau$ is the confidence threshold, and $\gamma$ controls the steepness of the sigmoid curve. This transformation suppresses noise below a defined confidence threshold while preserving high-confidence detections. By modulating the steepness of the transition around the threshold $\tau$, the function produces a smooth, interpretable confidence curve. The output yields a minimal satisfying interval (e.g., frames $615 \rightarrow 820$) that captures query-relevant events with high temporal precision. 
    
    \paragraph{Temporal Extension.}
    We extend the identified interval to ensure a complete temporal context for robust question answering (\texttt{L13}). The context window extension process extends the interval forward and backward by a fixed or learned temporal window $\alpha, \beta$, ensuring that sufficient temporal context is preserved for comprehensive understanding. We leverage the VLM to account for event lead-up and aftermath periods following temporal keywords such as ``before" and ``after" in the query $\textbf{q}$. The final extended window becomes the filtered input to the VLM for question answering, providing a temporally focused, yet contextually complete video segment for optimal performance. The temporal extension is formalized in \texttt{L14}.

    \paragraph{Vision Language Model Answering.}
        \label{subsec:vlm_answering}
        The final stage uses this temporally filtered video segment as context to generate accurate responses through VLM prompting.

\section{Experimental Setup}
    During automaton construction, we use InternVL2-8B \citep{internvl2024v2} as the VLM to generate frame-level detections. The videos are sampled at 3 frames per second. To translate natural language questions into TL specifications, we use the GPT-o1-mini reasoning model \citep{achiam2023gpt} with a few-shot prompting strategy and a carefully designed system prompt to capture common temporal structures and patterns. We apply Stormpy \citep{storm, stormpy}, a probabilistic model checker, to identify video segments that satisfy these TL specifications. We present the results of \nsvqa\ as a plug-and-play module with various VLMs in \Cref{sec:results}. 

    \paragraph{Datasets.}
    For empirical evaluation of \nsvqa\, we use two LVQA datasets: LongVideoBench \citep{wu2024longvideobench} and CinePile \citep{rawal2024cinepile}. 
    LongVideoBench serves as our primary benchmark. It features videos up to 60 minutes long across diverse domains such as instructional content, egocentric vlogs, and educational streams. CinePile is used as a secondary dataset to demonstrate the performance of \nsvqa\ on long-form movie scenes that require reasoning about narrative structure, character interaction, and scene progression. For both datasets, we focus on \textit{subsets} of question-video pairs that require \textit{temporal ordering}, \textit{causal inference}, and \textit{compositional logic}, filtering out simple examples based solely on static visual recognition. In LongVideoBench, these reasoning-oriented queries correspond to four categories: \textbf{\texttt{T3E}} (event before/after text), \textbf{\texttt{E3E}} (event before/after event), \textbf{\texttt{T3O}} (object before/after text), and \textbf{\texttt{O3O}} (object before/after object). To streamline the evaluation, we burn the provided subtitles into the video, allowing \nsvqa\ and consequently any VLM to directly infer and answer questions involving dialogue without requiring separate time-aligned subtitle prompts. 

    \paragraph{Evaluation Metrics and Benchmarking.}
    We evaluate \nsvqa\ on multiple-choice LVQA tasks using \textit{accuracy} as the primary metric across both LongVideoBench and CinePile. Since \nsvqa\ serves as a plug-and-play key-segment retrieval layer, we apply it modularly on top of several state-of-the-art VLMs, including Qwen2.5 \citep{bai2025qwen2}, GPT-4o \citep{achiam2023gpt}, and LLaVA-OneVision \citep{li2024llava}. For each model, we report performance with and without our retrieval pipeline to measure the impact of symbolic retrieval.
    In addition to these baselines, we compare against recent structured LVQA methods such as T*, VideoTree, and LVNet \citep{ye2025re, wang2025videotree, park2024too}, as discussed in \Cref{sec:related-works}. All benchmarks are conducted using the \texttt{lmms-eval} library \citep{zhang2024lmms}, which provides a unified framework for evaluating LVQA models across datasets. 


\section{Results}
\label{sec:results}

    Building on our experimental setup, we now turn to empirical evaluation. Our goal is to answer the following key questions regarding neuro-symbolic LVQA:
    
    \begin{enumerate}
        \item To what extent does \nsvqa\ improve VQA performance on significantly longer videos?
        \item Do the frames selected by \nsvqa\ add more relevant context that improves VQA performance?
        \item How crucial is the neurosymbolic reasoning component in \nsvqa\ for effective long video understanding?
    \end{enumerate}

    \subsection{Neuro-Symbolic LVQA}
\begin{figure*}[t]
        \centering
        \includegraphics[width=\linewidth, clip, trim=0cm 0.8cm 2.2cm 0cm]{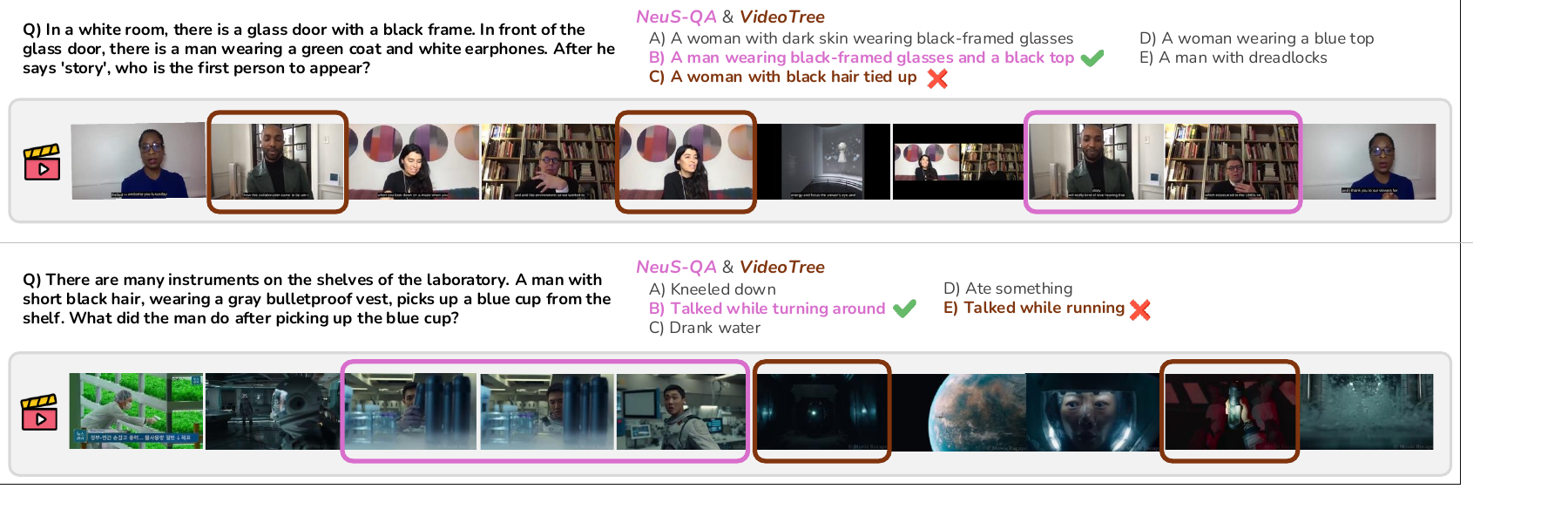}
        \caption{\textbf{Qualitative Examples from the \nsvqa\ pipeline.} Compared to other structured reasoning frameworks, \nsvqa\ more accurately identifies the correct frames of interest in a long video.}
        \label{fig:qualitative-results}
    \end{figure*}

        \paragraph{Evaluations on LongVideoBench.}
        We evaluate \nsvqa\ on LongVideoBench by comparing it against both VLMs and structured reasoning baselines. As shown in \Cref{tab:maintable}, \nsvqa\ improves VQA accuracy across all question types. Paired with Qwen2.5-VL, it achieves $60.09\%$ overall accuracy---nearly $10\%$ over the base model and outperforms all structured reasoning baselines, including VideoTree and LVNet. Qualitative examples in \Cref{fig:qualitative-results} further illustrate \nsvqa's ability to identify the correct frames of interest. These results highlight the benefits of symbolic temporal reasoning for long video understanding and demonstrate that dynamically selecting video segments based on the initial query is more effective than fixed or heuristic approaches.

\begin{figure}
        \centering
        \includegraphics[width=0.9\linewidth]{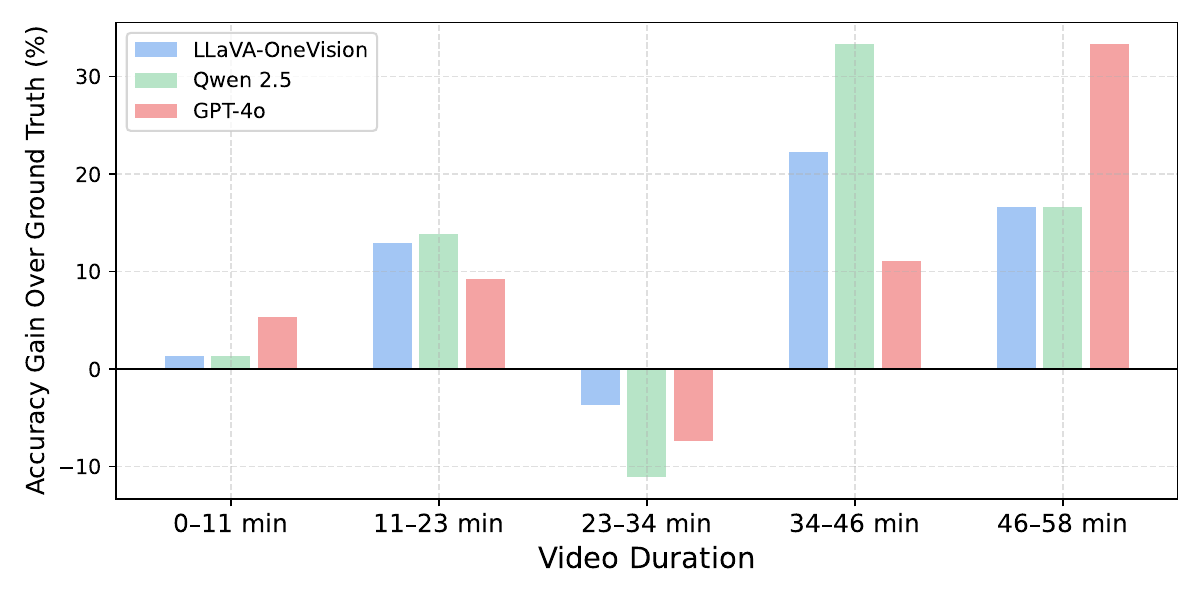}
        \caption{\textbf{Accuracy (\%) gains of \nsvqa\ over ground truth frame annotations.} We visualize the per-model improvement of \nsvqa\ in terms of absolute accuracy difference relative to ground truth annotations across different video durations. Positive values indicate that \nsvqa\ outperforms ground truth frame annotations.}
        \label{fig:gt-evals}
    \end{figure}

        We also assess scalability with increasing video duration as shown in \Cref{fig:gt-evals}, performance remains stable or improves, whereas prompting-based methods show substantial degradation. Notably, our method even outperforms human-annotated ground truth segments, achieving $59\%$ vs. 52\% when both are used as input to the same VLM. We attribute this to the automaton's ability to isolate not only the direct answer span but also semantically and temporally relevant context (e.g., events before/after).

        \paragraph{Performance on CinePile.}
         We evaluate \nsvqa\ on CinePile, a long-video benchmark designed for narrative-driven understanding. Unlike instructional or surveillance-style datasets, CinePile emphasizes abstract, story-based reasoning. As shown in Table~\ref{tab:cinepile-evals}, \nsvqa\ achieves 53.66\% accuracy when paired with Qwen2.5-VL, yielding notable gains over zero-shot prompting alone. These results indicate that our method generalizes effectively to more complex video domains, demonstrating its ability to identify contextually relevant content even without task-specific tuning.

    \begin{table}[t]
        \setlength{\tabcolsep}{1mm}
        \centering
        \resizebox{0.9\linewidth}{!}{%
            \begin{tabular}{clc}
                \toprule
                \multicolumn{2}{c}{\textbf{Strategy}} & \textbf{Overall ($\uparrow$)} \\
                \midrule
                \multirow{5}{*}{\shortstack[c]{Vision Foundation\\Models}}
                    & Qwen2.5-VL-7B-Instruct & 50.73 \\
                    & GPT-4o-2024-08-06 & 50.24 \\
                    & Llama-3.2-11B-Vision-Instruct & 20.98 \\
                    & LLaVA-OneVision-Qwen2-7B & 31.71 \\
                    & Idefics2-8B & 25.85 \\
                \midrule
                \multirow{1}{*}{\shortstack[c]{\nsvqa\ (Ours)}} 
                    & Qwen2.5-VL-7B-Instruct & \textbf{53.66} \\
                \bottomrule
            \end{tabular}
        }
        \caption{\textbf{Generalization to Narrative Video.} \nsvqa\ achieves the highest accuracy (\%) on CinePile, demonstrating its effectiveness in abstract, story-driven video domains.}
        \label{tab:cinepile-evals}
        \vspace{-1em}
    \end{table}

    \subsection{Ablation on \nsvqa\ Architectural Decisions}

        \paragraph{Frame Presentation Strategy.}
         Frame presentation affects how much context the model can leverage when answering questions. We compare the default \nsvqa\ approach of trimming the video to only the most relevant segments against alternative approaches that retain full temporal context while varying frame sampling density, sampling densely within relevant intervals and sparsely elsewhere. As shown in Table~\ref{tab:frame_presentation_ablation}, these adaptive strategies recover some incorrect answers but result in lower overall accuracy. This suggests that while context retention offers some benefit, tightly focusing on relevant segments remains more effective.

        \paragraph{Impact of Automata VLM Choice.}
        The choice of vision-language model used to identify atomic propositions during automaton construction directly affects the quality of the extracted frames of interest. We compare InternVL2-8B and Qwen2.5-VL for this step, keeping the downstream VQA model fixed. As shown in Table~\ref{tab:vlm_choice_ablation}, InternVL2-8B leads to higher VQA accuracy, even though both models produce similar median clip lengths. This suggests that stronger VLMs that more accurately identify propositions yield more faithful automatons, improving the precision and relevance of the frames selected for reasoning.

    \begin{table}[t]
        \centering
        \resizebox{0.95\linewidth}{!}{
            \begin{tabular}{ccccc}
                \toprule
                \textbf{Choice of VLM} & \textbf{Median Clip Size (s)} & \textbf{Accuracy ($\uparrow$)} \\
                \midrule
                Qwen2.5-VL-7B-Instruct & 40 & 52.63 \\
                InternVL2-8B & 40 & \textbf{58.33} \\
                \bottomrule
            \end{tabular}
        }
        \caption{\textbf{Impact of VLM backbone choice on accuracy (\%).} Using InternVL2-8B for automaton construction results in the highest accuracy when evaluated with GPT-4o.}
        \label{tab:vlm_choice_ablation}
    \end{table}

    \begin{table}[t]
        \centering
        \resizebox{\linewidth}{!}{
            \begin{tabular}{lccc}
                \toprule
                \textbf{Strategy} & \textbf{Accuracy} ($\uparrow$) & \textbf{Improvements} ($\uparrow$) & \textbf{Regressions} ($\downarrow$)\\
                \midrule
                Original (Full) & 49.56 & -- & --\\
                \nsvqa\ (Trim) & \textbf{58.33} & \textbf{18.42} & 9.65 \\
                Dynamic Scaling & 53.51 & 12.28 & 8.33 \\
                Speed Bias & 52.19 &10.53 & \textbf{7.90} \\
                \bottomrule
            \end{tabular}
        }
        \caption{\textbf{Impact of frame sampling strategy.} Trimming to relevant segments with \nsvqa\ yields the highest accuracy (+9.77\%) over the full-video baseline, driven by a substantial increase in correct predictions (+18.42\%) and a moderate rise in regressions (+9.65\%). Dynamic scaling and speed bias strategies offer intermediate trade-offs, illustrating the balance between context retention and focus.}
        \vspace{-1em}
        \label{tab:frame_presentation_ablation}
    \end{table}

    \subsection{How Important is Neuro-Symbolic Feedback on Frame Selection?}
    \label{sec:how-imp-neus}
        To isolate the impact of formal reasoning, we evaluate a simplified variant of \nsvqa\ that skips temporal logic and automaton construction. Instead, it relies \textit{solely} on prompting a VLM to judge whether a segment is relevant to the question. As shown in \Cref{tab:no_nsvs}, this approach underperforms significantly: without structured semantics or temporal constraints, \textbf{frame selection becomes noisy} and accuracy drops. These findings underscore the importance of neuro-symbolic feedback in guiding segment retrieval with interpretable, temporally grounded structure. 

    \begin{table}[t]
        \centering
        \resizebox{0.9\linewidth}{!}{
            \begin{tabular}{lcc}
                \toprule
                \textbf{Metric} & \textbf{\nsvqa\ (Ours)} & \textbf{VLM-only} \\
                \midrule
                Accuracy (0s, 600s] & \textbf{61.33} & 42.67 \\
                Accuracy (600s, 3600s] & \textbf{56.86} & 38.82 \\
                Overall Accuracy ($\uparrow$) & \textbf{58.33} & 40.09 \\
                \bottomrule
            \end{tabular}
        }
        \caption{\textbf{Effect of neuro-symbolic feedback on frame selection.} Replacing temporal logic and automaton construction with direct VLM queries significantly degrades accuracy (\%) across both short and long video segments.}
        \vspace{-1em}
        \label{tab:no_nsvs}
    \end{table}


\section{Discussion}
    \textbf{Applicability Beyond Explicit Temporal Prompts.}
    While \nsvqa\ is designed to handle complex temporal queries, its applicability extends even to cases where temporal structure is \underline{implicit}. Since we use an LLM to translate questions into temporal logic, it can infer event order from linguistic cues, even if the prompt \textit{does not} contain explicit temporal operators such as ``before" or ``after." In simpler or ambiguous queries, the system effectively treats the entire prompt as a single proposition, reducing the reasoning task to identifying the most relevant clip. In such degenerate cases, performance aligns with standard VLM-based frame retrieval approaches, as shown in Section~\ref{sec:how-imp-neus}. In Table~\ref{tab:maintable}, we focus on four temporally compositional categories from LongVideoBench where our approach offers the greatest benefits. Results for other categories (see Appendix) also show improvements due to our logic-based retrieval, though gains are less pronounced due to the reduced temporal complexity of those queries.

    \paragraph{Choice of Temporal Logic Formalism.}
    Another natural question is: ``Why temporal logic?" Is it the only or best choice for structuring complex queries? We adopt TL because it strikes a balance between expressivity and compatibility with existing formal verification tools such as Storm and CARL \citep{storm, carl}. TL captures temporal constructs like ordering, simultaneity, and eventuality while remaining amenable to model checking. However, \nsvqa\ is \textbf{logic-agnostic} too: alternatives like Linear Temporal Logic (LTL), Signal Temporal Logic (STL), or \textit{any} domain-specific structured representation could be substituted, as long as they provide a mechanism to create and verify the logical conditions of a query. 
    
    This flexibility makes \nsvqa\ a \textbf{general framework} for interpretable long-video reasoning. Our contribution is \textbf{not} about choosing these \textit{design} and \textit{architectural} choices, but about enabling \textbf{symbolic alignment} between video content and query structure as a strategy for aiding LVQA. 

    \paragraph{Limitations and Failure Modes.}
    A key limitation of \nsvqa\ lies in the coupling between its ``neuro" and ``symbolic" components. Although the symbolic verifier is \textit{exact} by construction (guaranteeing correctness once propositions are grounded), the quality of the output depends heavily on the VLM's ability to \textit{detect} relevant events. In cases with multiple complex propositions, a single missed detection can cause the TL specification to be evaluated as unsatisfied, leading to false negatives during clip retrieval. These failure cases are most common when visual cues are subtle, occluded, or appear briefly in the video.

    \paragraph{Future Work.}
    Looking forward, we see opportunities for extending \nsvqa\ beyond one-shot retrieval and VQA. One direction is to explore \textit{agentic} neuro-symbolic systems that reason over long videos \underline{iteratively} using memory, planning, and intermediate logic representations to decompose and refine queries. Additionally, while our video automaton representation scales well with video length, constructing it remains computationally expensive. In theory, this cost could be amortized by pre-building a general-purpose video automaton and applying it to multiple downstream queries in a coarse-to-fine retrieval loop. 

    \paragraph{Broader Implications.}
    While \nsvqa\ achieves strong improvements and sets new state-of-the-art results on LVQA benchmarks (Table~\ref{tab:maintable}), overall scores including those from VLMs prompted with ground truth clips (Figure~\ref{fig:gt-evals}) remain modest. This highlights the broader limitations of current perception models. As VLMs \textit{continue to improve}, the quality of both frame-level grounding and automaton construction \textit{will scale accordingly}.

\section{Conclusion}
    We introduce \nsvqa, a plug-and-play, training-free neuro-symbolic framework for long-form video question answering. By translating natural language queries into temporal logic and verifying them over a video automaton, \nsvqa\ isolates segments semantically and temporally aligned with the question before querying a VLM. Our results show strong gains over standard prompting and recent retrieval-based methods \citep{wang2025videotree, park2024too}, especially on temporally complex benchmarks like LongVideoBench and CinePile \citep{wu2024longvideobench, rawal2024cinepile}. As VLMs improve, the fidelity of our symbolic alignment will scale in tandem. \nsvqa\ is a general framework for structured, interpretable long-video reasoning. The key insight lies not in any particular system design or architecture, but symbolically aligning video content with queries as an effective and principled approach to LVQA.

\section*{Acknowledgments}
This material is based upon work supported in part by the Office of Naval Research (ONR) under Grant No. N00014-22-1-2254. Additionally, this work was supported by the Defense Advanced Research Projects Agency (DARPA) contract DARPA ANSR: RTX CW2231110. Approved for Public Release, Distribution Unlimited.

\bibliography{aaai2026}

\end{document}

%% file: preamble.tex
\usepackage{xcolor}
\usepackage{lipsum} 
\usepackage{pifont} 
\usepackage{multirow}
\usepackage{booktabs}
\usepackage{colortbl}
\usepackage{amsmath}
\usepackage{amssymb}
\usepackage[linesnumbered, ruled,vlined]{algorithm2e}
\usepackage{cleveref} 
\usepackage{subcaption}
\usepackage{placeins}  

\makeatletter
\newcommand{\removelatexerror}{\let\@latex@error\@gobble}
\makeatother

\usepackage{ifthen} 
\usepackage{etoolbox} 
\newboolean{showcomments}
\setboolean{showcomments}{true} 
\newrobustcmd{\sps}[1]{\ifthenelse{\boolean{showcomments}}{\textcolor{orange}{(\textbf{SP}: #1)}}{}}
\newrobustcmd{\mk}[1]{\ifthenelse{\boolean{showcomments}}{\textcolor{green}{(\textbf{Minkyu}: #1)}}{}}
\newrobustcmd{\sas}[1]{\ifthenelse{\boolean{showcomments}}{\textcolor{purple}{(\textbf{Sahil}: #1)}}{}}
\newrobustcmd{\merc}[1]{\ifthenelse{\boolean{showcomments}}{\textcolor{blue}{(\textbf{HG}: #1)}}{}}



%% file: macros.tex
\newcommand{\propset}{\mathcal{P}}
\newcommand{\prop}{p}
\newcommand{\spec}{\varphi}
\newcommand{\video}{\mathcal{V}}
\newcommand{\auto}{\mathcal{A}}
\newcommand{\va}{\auto_\video}

\newcommand{\andlogic}{$\wedge$}
\newcommand{\orlogic}{$\vee$}
\newcommand{\notlogic}{$\neg$}
\newcommand{\imply}{$\implies$}

\newcommand{\always}{$\Box$}
\newcommand{\eventually}{$\diamondsuit$}
\newcommand{\ournext}{$\mathsf{X}$}
\newcommand{\until}{$\mathsf{U}$}

\newcommand{\nsvqa}{\textit{NeuS-QA}}
\newcommand{\puls}{\mathcal{L}_\texttt{Q2TL}}